\documentclass[letterpaper, 10 pt, conference]{ieeeconf}  

\IEEEoverridecommandlockouts                              

\usepackage{hyperref}
\usepackage{listings}
\usepackage{siunitx}
\usepackage{multirow}
\usepackage[pdftex]{graphicx}
\usepackage[labelformat=simple]{subcaption}
\usepackage{xcolor}
\usepackage{cite}
\usepackage{paralist}
\usepackage{nicefrac}
\usepackage[ruled,vlined,linesnumbered]{algorithm2e}

\DeclareCaptionLabelSeparator{periodspace}{.\quad}
\let\labelindent\relax
\usepackage{enumitem}
\newcommand{\vrep}{{\sc v-rep}}

\title{\LARGE \bf
Path Planning With Kinematic Constraints For Robot Groups
}

\author{Wolfgang H\"onig, T.~K.~Satish Kumar, Liron Cohen, Hang Ma, Sven Koenig, Nora Ayanian
\thanks{The authors are with the Department of Computer Science at the University of Southern California, USA.
        {\tt\small whoenig@usc.edu, tkskwork@gmail.com, \{hangma,lcohen,skoenig,ayanian\}@usc.edu}.
Our research
    was supported by ARL under grant number W911NF-14-D-0005, ONR under grant
    numbers N00014-14-1-0734 and N00014-09-1-1031, NASA via Stinger Ghaffarian
    Technologies, and NSF under grant numbers 1409987 and 1319966.
        }
}

\begin{document}

\maketitle
\thispagestyle{empty}
\pagestyle{empty}

\begin{abstract}
Path planning for multiple robots is well studied in the AI and robotics communities.
For a given discretized environment, robots need to find collision-free paths to a set of specified goal locations. Robots can be fully anonymous, non-anonymous, or organized in groups.
Although powerful solvers for this abstract problem exist, they make simplifying assumptions by ignoring kinematic constraints, making it difficult to use the resulting plans on actual robots.
In this paper, we present a solution which takes kinematic constraints, such as maximum velocities, into account, while guaranteeing a user-specified minimum safety distance between robots.
We demonstrate our approach in simulation and on real robots in 2D and 3D environments.
\end{abstract}

\section{INTRODUCTION}

Path planning for multiple robots has many applications, including improving traffic at intersections, search and rescue, formation control, warehouse management, airport scheduling, and assembly planning.
There are two existing major approaches: The first one works in continuous environments and can take kinematic constraints into
account but does not perform well in highly cluttered, puzzle-like scenes, and the second one works in discrete environments
with artificial agents without motion constraints.

Hence, it is desirable to combine the
two approaches by providing a planner which can deal with
highly cluttered, puzzle-like scenes even under kinematic
constraints.
We tackle this challenge by introducing a postprocessing
step that works on the output of a discrete solver.
While the solver itself is allowed to make simplifying assumptions in order to run faster, our postprocessing step reinstates the adherence to real-world kinematic constraints.

Solvers from the AI community for \emph{Multi-Agent Path-Finding} (MAPF) problems and \emph{Target-Allocation and Path-Finding} (TAPF) problems are able to solve instances with hundreds of agents~\cite{ma2016b}.
For the MAPF problem, each agent has an assigned start- and goal location. The objective is to find a set of synchronized paths, one path per agent, such that each agent reaches its goal location without colliding with other agents while minimizing the number of actions required.
Each agent can either move to an adjacent location in one timestep or wait.
The TAPF problem is a generalization in which the agents are partitioned into $K$ groups.
A set of goal locations is assinged to each group. A solver allocates a specific
goal location to each agent and reports a set of synchronized paths for all agents.
If $K=1$, all agents are \emph{anonymous}, and, if $K$ equals the number of agents, TAPF is the same as MAPF. Therefore, we will focus on TAPF in the remainder of this paper.

Using a TAPF solution on real robots
has several limitations: (a) robots have kinematic constraints,
such as maximum velocities and accelerations; and (b) the
generated solution's timing is inflexible, necessitating costly replanning when robots execute the solution imperfectly.
A framework that does not explicitly
address faulty execution can cause undesirable robot-robot
collisions or repeated replanning.
To overcome these limitations while still being able to
make use of powerful solvers developed in the AI
community, we propose the use of a postprocessing step
based on the algorithmic framework of Simple Temporal
Networks (STNs).

\section{TAPF}

\begin{figure}[t]
    \begin{subfigure}[t]{0.23\textwidth}
        \centering
        \includegraphics[width=0.8\textwidth]{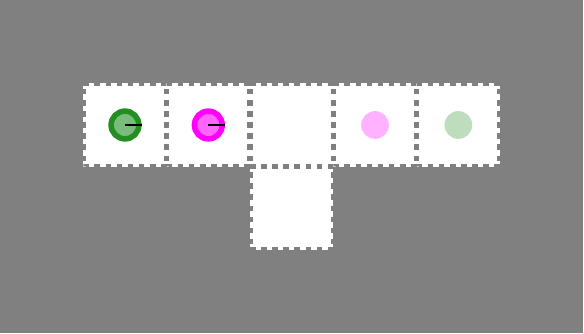}
        \caption{Example environment with two robots. The lighter circles on the right mark the goal locations.}
        \label{fig:example:env}
    \end{subfigure}
    \hfill
    \begin{subfigure}[t]{0.23\textwidth}
        \centering
        \includegraphics[width=0.8\textwidth]{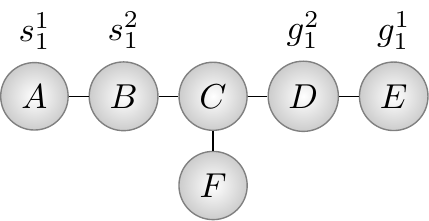}
        \caption{Graph representation of the example environment.}
        \label{fig:example:graph}
    \end{subfigure}
    \caption{Running TAPF example with 2 groups.}
    \vspace{-15pt}
    \label{fig:example}
\end{figure}

We are given an undirected graph $G_1=(V_1,E_1)$ and $K$ multi-agent groups $\{\text{group}_1,\text{group}_2,\ldots,\text{group}_K\}$, where $\text{group}_i$ consists of $K_i$ interchangeable robots $\{a_1^i,a_2^i,\ldots,a_{K_i}^i\}$ for all $i\in\{1,2,\ldots,K\}$.
Each robot $a_j^i$ has a unique start location $s_j^i\in V_1$, and the $i$-th group has a set of unique target locations $\{g_1^i, g_2^i,\ldots,g_{K_i}^i\}$.
A solution to the TAPF problem finds $K$ permutations, one for each group, to uniquely assign a target location to each robot and a collision-free path for each robot to navigate from its start location to the assigned goal location. A more rigorous mathematical description is given in~\cite{ma2016b}.

The \emph{makespan} is the total time until the last robot reached its goal location. A solution is optimal if the makespan is minimal.
For the case of a single group, the problem can be solved in polynomial time. However, in general, it is NP-hard to approximate an optimal solution within any constant factor less than $\nicefrac{4}{3}$~\cite{ma2016a}.

The Conflict-Based Min-Cost-Flow (CBM) algorithm uses a hierarchical approach, where the lower level uses a max-flow algorithm and the higher level uses a best-first search, which tries to resolve conflicts as they occur~\cite{ma2016b}.
It has been shown empirically that this method works well in warehouse domains with dozens of teams and hundreds of agents.

An example of a TAPF instance is shown in Fig.~\ref{fig:example}. Here, there are two groups, that each contain a single robot. The robots are holonomic and move in a discretized environment.
One optimal solution of the given TAPF problem is $\langle A,B,C,D,E\rangle$ for one robot and $\langle B,C,F,C,D\rangle$ for the other robot.

\section{STN}

\begin{figure}[t]
    \centering
    \includegraphics[width=0.49\textwidth]{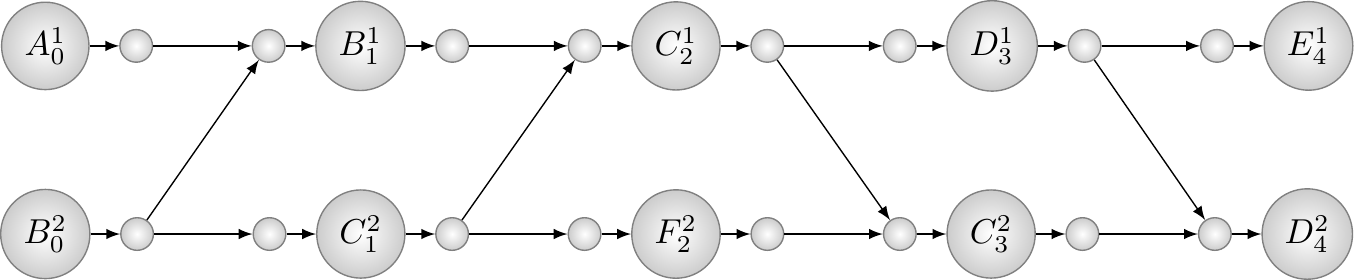}
    \caption{Temporal plan graph with safety markers.}
    \vspace{-15pt}
    \label{fig:tpg}
\end{figure}

The output of the TAPF solver is a set of synchronized paths, assuming that robots can traverse the unit-length edges in unit time.
We call the change of location of an agent \emph{event}.
In order to include kinematic constraints, such as maximum velocities of the different robots, we use a \emph{Temporal Plan Graph} (TPG).
The TPG captures the partial order of the events determined by the TAPF solver.
More formally, the TPG is an acyclic graph $G_2=(V_2,E_2)$, where each vertex $v\in V_2$ represents an event and each edge $\langle u,v\rangle\in E_2$ indicates that $u$ should be scheduled before $v$.

In the TAPF case, there are two kinds of temporal precedences. First, the precedence which captures the location visit order for each robot individually. Second, the TAPF solution synchronizes the paths between robots to avoid conflicts. This happens if robot 1 visits a location at timestep $t_1$ and robot 2 visits the same location at timestep $t_2>t_1$.

Moreover, we can add \emph{safety markers} as additional events to ensure a minimum safety distance between robots at any time. There are several methods to do so. An LP-solver can be used to optimize for maximum throughput or minimum makespan~\cite{hoenig-mapf}. If the desired safety distance divides the edge length, the problem can be solved with a user-specified edge-length in strongly polynomial time by using a shortest path algorithm~\cite{hoenig-tapf}.
One example of a TPG with safety markers is shown in Fig.~\ref{fig:tpg}.

The TPG captures only the necessary partial order on events.
In order to include kinematic constraints such as maximum velocities for robots or certain parts of the environment, we can extend it to a \emph{simple temporal network} (STN).
An STN can be encoded as directed graph $G_3=(V_3,E_3)$, where $V_3=\{X_0,X_1,\ldots,X_N\}$ and $E_3$ are the sets of events and edges, respectively.
Each edge $e=\langle X_i, X_j\rangle\in E_3$ has lower and upper bounds $[LB(e), UB(e)]$, indicating that $X_j$ has to be scheduled between $LB(e)$ and $UB(e)$ time units after $X_i$.


\section{EXPERIMENTS}

We implement TAPF and MAPF solvers as well as various variants of the STN framework in C++.
For performance evaluation, we randomly generate $10\times 10\times 5$ maps with varying numbers of robots, groups, and obstacles and measure how long the TAPF solver takes to find an optimal solution.
For example, a scenario with $150$ obstacles and $100$ robots in $5$ groups can be solved in about \SI{5}{s} on commodity hardware.
The time the STN requires to compute a solution varies by method and desired safety distance. Interestingly, its runtime is smaller in the common case of a large safety distance, because fewer markers are required in that case. In a warehouse-like domain with $100$ robots, it takes \SI{4}{s} to compute a solution.

\begin{figure}[t]
        \centering
        \includegraphics[width=.45\textwidth]{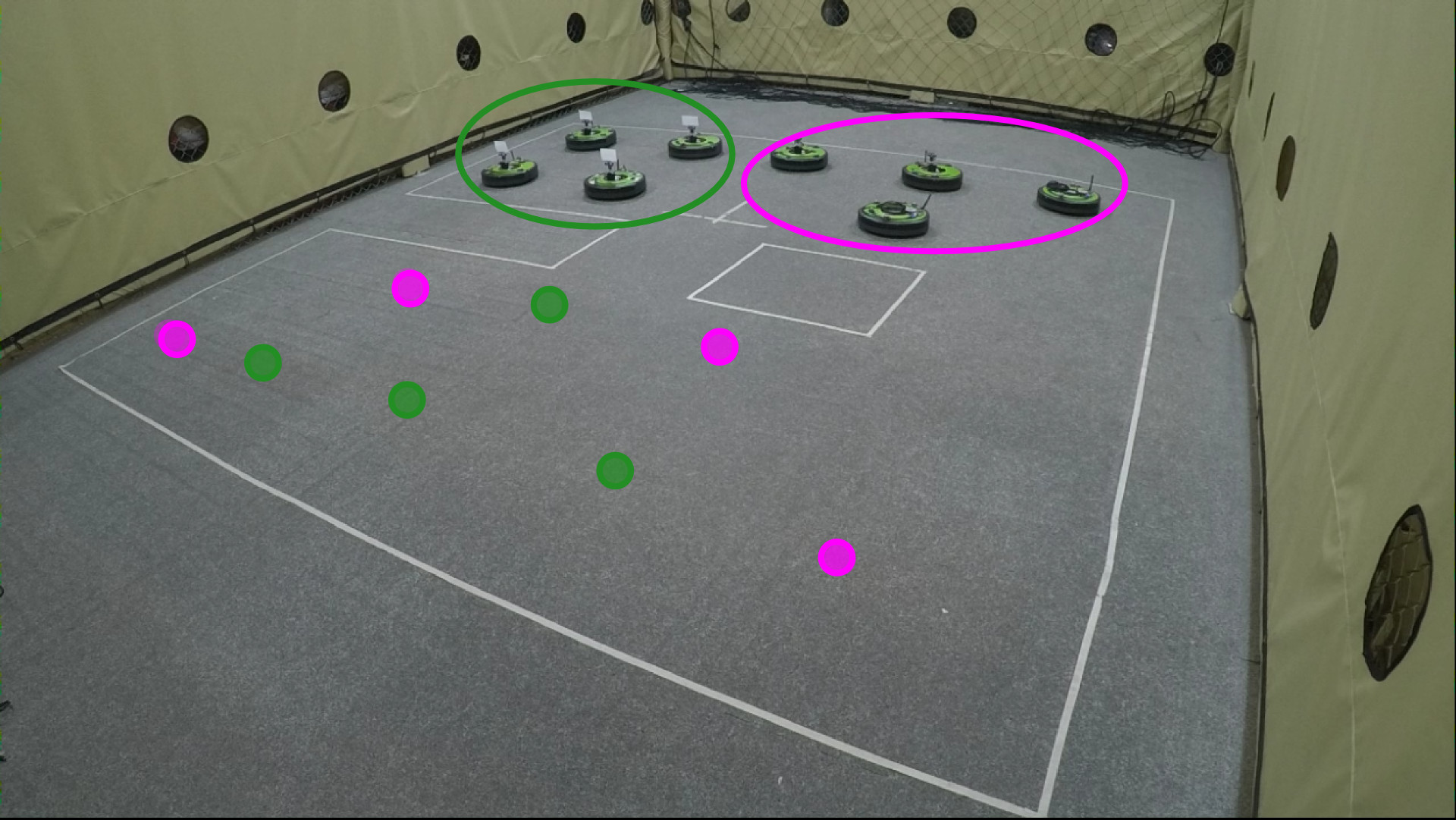}
    \caption{Examples of the experiments on robots. There are two groups of four robots each trying to change formation in an environment with obstacles.}
    \vspace{-15pt}
    \label{fig:exp}
\end{figure}

We verified our approach in simulation using the \vrep{} robotics simulator for differential drive robots, $6$-legged robots, and quadcopters. Furthermore, we performed experiments with up to $8$ iRobot Create2 robots, verifying that actual robots are actually able to follow the computed trajectories. An example is shown in Fig.~\ref{fig:exp}.

\section{CONCLUSION}

We presented an approach for using powerful solvers from the AI community for multi-agent path finding on actual robots, which obey kinematic constraints. Although not shown here, our approach is optimal with respect to the makespan under some conditions. We demonstrate the applicability of our approach through simulation and on actual robots.
In the future, we would like to test our framework on physical quadcopters and add an execution monitoring framework which makes use of the ``slack'' of the STN to avoid costly replanning in case of inaccurate execution.

\bibliographystyle{IEEEtran}
\bibliography{IEEEabrv,references}

\end{document}